%% file: main.tex
\documentclass[review]{llncs}
\usepackage[T1]{fontenc}
\usepackage{graphicx}
\usepackage{cite}
\usepackage{amsmath,amssymb,amsfonts}
\usepackage{algorithm}
\usepackage[noend]{algorithmic}
\usepackage{graphicx}
\usepackage{textcomp}
\usepackage{xcolor}
\usepackage{appendix}
\usepackage{xspace}
\usepackage{times}
\usepackage{soul}
\usepackage{tikz}

\newcommand{\scenic}{{\sc Scenic}\xspace}

\begin{document}
\title{Querying Labeled Time Series Data \\with Scenario Programs}
\author{Edward Kim\textsuperscript{*\textdagger}, Devan Shanker\textsuperscript{*\textdagger}, Varun Bharadwaj\textsuperscript{\textdagger}, Hongbeen Park\textsuperscript{\textdagger\textdagger}, Jinkyu Kim\textsuperscript{\textdagger\textdagger}, Hazem Torfah\textsuperscript{$\ddagger$}, Daniel J. Fremont\textsuperscript{$\ddagger\ddagger$}, Sanjit A. Seshia\textsuperscript{\textdagger}}
\institute{\textsuperscript{\textdagger} University of California, Berkeley, \textsuperscript{\textdagger\textdagger} Korea University, South Korea,\\
\textsuperscript{$\ddagger$} Chalmers University of Technology and University of Gothenburg, Sweden,\\
\textsuperscript{$\ddagger\ddagger$} University of California, Santa Cruz,\\
(* first co-authors contributed equally)
}
% %

\maketitle              % typeset the header of the contribution
% %
\begin{abstract}
Simulation-based testing has become a crucial complement to road testing for ensuring the safety of cyber-physical systems (CPS). As a result, significant research efforts have been directed toward identifying failure scenarios within simulation environments. However, a critical question remains: are the AV failure scenarios discovered in simulation \textit{relevant} to real-world systems—specifically, are they reproducible on actual systems? The sim-to-real gap caused by differences between simulated and real sensor data means that failure scenarios identified in simulation might either be artifacts of synthetic sensor data or actual issues that also occur with real sensor data. To address this, an effective approach to validating simulated failure scenarios is to locate occurrences of these scenarios within real-world datasets and verify whether the failure persists on the datasets. To this end, we introduce a formal definition of how labeled time series sensor data can match an abstract scenario, represented as a scenario program using the \scenic{} probabilistic programming language. We present a querying algorithm that, given a scenario program and a labeled dataset, identifies the subset of data that matches the specified scenario. Our experiment shows that our algorithm is more accurate and orders of magnitude faster in querying scenarios than the state-of-the-art commercial vision large language models, and can scale with the duration of queried time series data.

\end{abstract}

\section{Introduction}
\input{sections/introduction}\label{sec:intro_related}

% \section{Related Work}
% \input{sections/related_work}

% \section{Background}\label{sec:background}
% \input{sections/background}

\section{Background and Overview}\label{sec:overview}
\input{sections/overview}

\section{Problem Formulation}\label{sec:problem}
\input{sections/problem}

\section{Methodology}
\label{sec:method}
\input{sections/methodology}

\section{Experiment}
\label{sec:expt}
\input{sections/experiment}

% \section{Discussion}
% \input{sections/discussion}

% \section{Limitations}
% \input{sections/limitations}

\section{Conclusion}
\input{sections/conclusion}

\bibliographystyle{IEEEtran}
\bibliography{reference}

\appendix
\input{sections/appendix}

\end{document}

%% file: sections/introduction.tex
Simulation-based testing and verification has become an integral part of the design of safety-critical artificial intelligence (AI)-enabled cyber-physical systems (CPS) such as self-driving cars. There is a growing body of literature proposing algorithms which automate search for system failure inducing scenarios in simulation~\cite{dreossi-nfm17, verifai, adaptive-icra, advsim, kim-cvpr20}. These simulation-based testing techniques contribute to scalable assessments of autonomous systems without the risk of injuring other people who may interact with these systems.

However, given that the primary objective of simulation-based testing is to improve the performance and safety of CPS in reality (not in simulation), it is pertinent to ask: \textit{``are failure scenarios identified in simulation reproducible in reality?''} Due to discrepancies between \textit{synthetic} versus \textit{real} sensor data, the systems under evaluation may behave differently. Mathematically characterizing such discrepancies is very difficult, and it is not well understood how these discrepancies affect the behaviors of AI systems comprising deep neural networks. This problem is often referred to as the \textit{sim-to-real validation problem}. As the result of this problem, the status quo is to physically reconstruct failure scenarios and test the systems in reality, e.g., track testing for self-driving (e.g.,~\cite{fremont-itsc20}), which is labor intensive and not a scalable validation method.

%In this paper, we propose a \textit{data-driven} approach for the sim-to-real validation problem. 
In this paper, we propose a novel approach for sim-to-real validation that {\em automatically validates real data against formal models}. 
Formal approaches to world (environment) modeling are crucial to certifying the safety of AI-based autonomy~\cite{seshia-cacm22a}. 
Our work leverages current trends where (1) large amounts of \textit{real} data~\cite{nuscenes, kitti, waymodataset} are being collected and labeled to train systems, and (2) formal scenario modeling languages are increasingly adopted by the research communities working on autonomous CPS and robotics~\cite{scenic,scenic-journal,msdl}. Based on these trends, we propose to address the problem by \textit{querying} the dataset to find real-world instances of the scenario. Specifically, given (i) a formal model of a candidate failure scenario and (ii) a labeled time series dataset, we develop a query algorithm that outputs the subset of the labels (and, thus, the real sensor data) that \textit{matches} the scenario. Then, the system can be evaluated on the matching real data for validation. If no data is returned from the query, this provides a valuable indication that the failure scenario is not represented in the training dataset.

To precisely define the query problem and prove a correctness guarantee for our query algorithm, we formally define what it means for labeled time series data, or \textit{label trace}, to \textit{match} a scenario modeled in the probabilistic programming language, \scenic{}~\cite{scenic,scenic-journal}. 
A \scenic{} program represents a \textit{distribution} over the initial conditions and behaviors of objects. Given an input trace of observations of the world, the program defines a distribution over traces of discrete actions for each object. Given a label trace providing an observation sequence and the corresponding actions of objects, we define a label trace to \textit{match} a scenario if (1) the initial scene, i.e. the first element of the observation trace, is in the initial state distribution of the program, and (2) the action trace is in the distribution of the output traces generated by the program given the input trace.

\textbf{Related Work.} 
In database systems, numerous video retrieval approaches have been proposed such as BlazeIt~\cite{blazeit}, VisualWorldDB~\cite{visualworlddb}, SVQL~\cite{svql}, and ExSample~\cite{exsample}. These are built using an extension of structured query language (SQL)~\cite{sql}, which severely limits the expressiveness of queries specifying, for instance, only the presence of particular classes of objects, e.g. retrieve videos which contain a motorcycle and a truck. Spatialyze~\cite{spatialyze} and our prior work~\cite{iccps} provide expressive query formalisms, yet can only specify \textit{static} conditions involving objects positions and orientations with respect to road structure; they cannot specify \textit{temporal} behaviors of objects. Conversely, while not as optimized for querying videos, the field of video understanding in computer vision backed by vision language models (VLMs)~\cite{vlm} can take very expressive queries in natural language. However, natural language can be an ambiguous way to model scenarios, and VLMs currently have limited performance and no guarantees on accuracy. By contrast, we use \scenic{}~\cite{scenic-journal} as a query formalism to expressively model scenarios with precise semantics, and provide a query algorithm with a correctness guarantee. In summary, in this paper, we contribute:
\begin{enumerate}
    \item A novel problem formulation of {\em querying a labeled time series dataset against a formal scenario model}, which is applicable to sensor data of any type (e.g., RGB, LiDAR, radar, etc.) assuming the data is appropriately labeled (Sec.~\ref{sec:problem}).
    \item A {\em sound query algorithm} for a fragment of \scenic{} which can be used to validate candidate failure scenarios for diverse perception, behavior prediction, and planning tasks (Sec.~\ref{sec:method}). 
    \item Experiments showing that the algorithm is {\em more accurate and orders of magnitude faster} than the state-of-the-art VLMs and can scale linearly to video duration (Sec.~\ref{sec:expt}).
\end{enumerate}

%% file: sections/overview.tex
\begin{figure}[tb]
    \centering
    \includegraphics[width=0.7\linewidth]{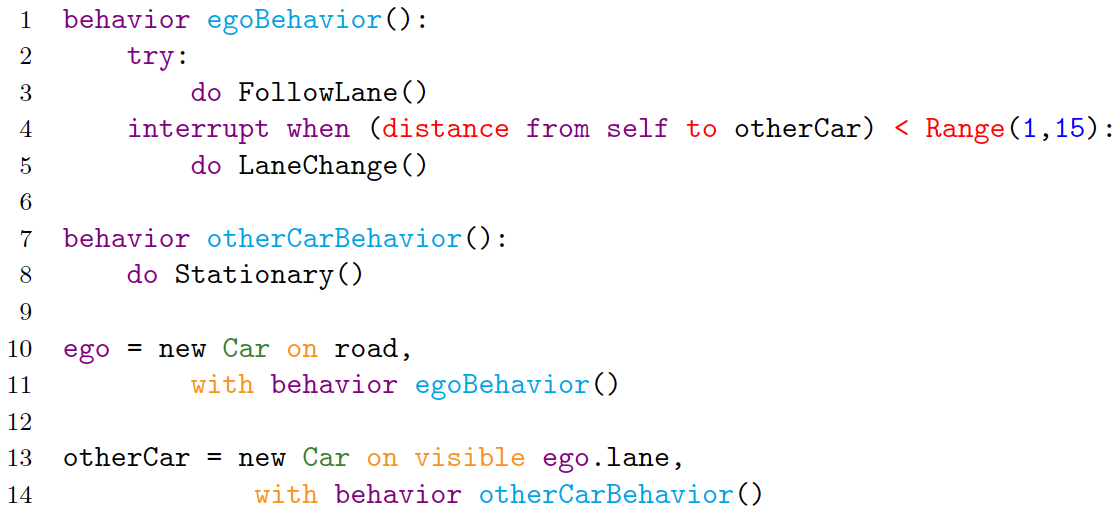}
    \caption{A \scenic{} program modeling an ego car making a lane change to avoid a car in front.}
    \label{fig:scenic-program}
\end{figure}

In this section, we provide context and examples to facilitate understanding the query problem formulation in Section~\ref{sec:problem}. Given a scenario description and labeled time series sensor data, our query problem is to determine whether the data \textit{matches} the scenario description. For brevity, and since video is the main type of time series data we use in this paper, we refer to time series sensor data as a \textit{video} from now on. \\
\indent\textbf{Query Language.} We adopt \scenic{}~\cite{scenic,scenic-journal} as our query language to formally model scenarios of interest. \scenic{} is a probabilistic programming language designed to model and generate scenarios in simulation. A \scenic{} program defines a set of objects, a distribution over their initial configuration, or \emph{scene} (i.e., position, orientation, etc.), and their probabilistic behaviors. The \scenic{} program in Fig.~\ref{fig:scenic-program} models a scenario where \texttt{ego} car follows a lane and makes a lane change to avoid \texttt{otherCar} that is stationary in front. Lines 10 and 13 define a distribution of initial conditions, \texttt{ego} and \texttt{otherCar}. For instance, in line 13, \texttt{otherCar} is uniformly randomly sampled on the lane of \texttt{ego}, which it intersects with \texttt{ego}'s view cone. Furthermore, \texttt{ego} is assigned a probabilistic behavior called \texttt{egoBehavior} which is defined on lines 1-5 using \texttt{try/interrupt} statement. By default, \texttt{ego} follows the lane it is on, but changes lane if its distance to the other car is within a distance uniformly randomly sampled from a range of 1 to 15 meters. Note that a behavior can be hierarchically defined using a set of \textit{primitive} behaviors, which in this case include ``follow lane'' and ``lane change.'' 

\indent\textbf{Semantics of the Query Language.} To define what it means for a video to match a query expressed in \scenic{}, we need to understand its semantics. A \scenic{} program first generates a scene defining the initial configuration of objects (line 10 and 13); the behaviors in the program then define how the objects behave as a function of the state of the world (e.g., their own positions and those of other objects).\\
\indent In particular, we view a behavior as taking as input a sequence of observations of the environment, forming an \textit{input trace}, and outputting a series of primitive behaviors, the corresponding \textit{output trace}.
Each element of an input trace is an assignment of concrete values to semantic features, such as objects' positions, orientations, and lanes occupied by objects. For example, given a finite input trace of three timesteps consisting of semantic feature values, e.g. positions and orientations of objects, the \texttt{egoBehavior} in Figure~\ref{fig:scenic-program} can generate a \textit{set} of output traces, e.g. $\{\langle \texttt{FL,FL,FL}\rangle , \langle \texttt{FL,LC,FL}\rangle\}$ where \texttt{FL} and \texttt{LC} are abbreviations for the primitive behaviors \texttt{FollowLane} and \texttt{LaneChange} respectively. Note that, due to the probabilistic aspect of the \scenic{} behavior, the \texttt{ego} behavior outputs a \textit{set} of possible primitive behaviors at each timestep. For example, in line 4 of Figure~\ref{fig:scenic-program}, the interrupt condition for triggering a lane change is defined over a uniform random distribution over an interval of 1 to 15 meters, i.e. \texttt{Range(1,15)}. When \texttt{ego} and \texttt{otherCar} are between 1 and 15 meters apart, either \texttt{FollowLane} or \texttt{LaneChange} could occur depending on the random distance threshold chosen.
In such a case, both primitive behaviors are feasible. \\
\indent This semantics of a single behavior extends to the semantics of an entire \scenic{} program through synchronous composition: the behaviors of all objects in the program run in parallel, with each one choosing a set of primitive behaviors in each time step. In our example, the \textit{set} of output traces for the entire program corresponding to the 3-step trace above would be:$\{\{\texttt{ego}:\langle\texttt{FL,FL,FL}\rangle, \texttt{otherCar}:\langle \texttt{St,St,St}\rangle\}, \{\texttt{ego}:\langle\texttt{FL,LC,FL}\rangle\}, \texttt{otherCar}:\langle \texttt{St,St,St}\rangle\}\}$, where each element contains the set of primitive behaviors of each object at each time step and \texttt{St} abbreviates the \texttt{Stationary}.\\
\indent\textbf{Label Trace.} In our query problem, it is important to note that we query the time series \emph{labels} of the video, not the raw sensor data. We refer to time series labels of the video as a \textit{label trace}. We assume the labels contain information that, as we will see in the problem description below, serves as the input and the output traces of the \scenic{} program. Specifically, we assume that the labels include observations (such as positions, orientations, occupied lanes), and primitive behaviors of objects. These observations can be computed using localization algorithms like SLAM~\cite{slam} and behavior prediction algorithms~\cite{behavior-prediction}, as we demonstrate in our experiments. To account for uncertainty in the classification of primitive behaviors, we allow the label trace to specify a \textit{set} of possible behaviors at each time step: for example, when using a neural network to predict primitive behaviors, one could include in the trace all behaviors whose predicted confidence is above a threshold.\\
\indent\textbf{Query Problem.} Suppose a \scenic{} program $P$ and a label trace $\ell$ are given. The problem is to determine whether the observations in the label trace \textit{match} the scenario modeled in the program. Intuitively, we first check if the initial scene, the first element of  the input trace of the label trace, is consistent with the scenario, i.e. the initial scene is in the support of the initial distribution of the program. Then, we check if all the objects in the label trace behave consistently with the program, i.e., whether there is some output trace of the program which agrees with the label output trace.
We formally define the problem in the next section.\\
\indent\textbf{Query Examples.} We provide a few examples based on querying with the \scenic{} program $P$ in Figure~\ref{fig:scenic-program}. Suppose a (simplified) input trace $\sigma_{in}^\ell= \{\texttt{ego}: [(0,0), (0,1),\\ (1,2), (1,3)],$ $\texttt{otherCar}: [(0,7),(0,7),(0,7),(0,7)]\}$ from the label trace $\ell$, consisting of the positions of observed vehicles are given. Let the set of output traces generated by the \scenic{} program is
$\sigma_{OUT}^\mathit{P}= \{\{\texttt{ego}:\langle\texttt{FL,FL,LC,FL}\rangle,$ $ \texttt{otherCar}:\langle \texttt{St,St,St,St}\rangle\}, \{\texttt{ego}:\langle\texttt{FL,LC,LC,FL}\rangle,\texttt{otherCar}:\langle \texttt{St,St,}$ $\texttt{St,St}\rangle\}\}$. Assume that the initial scene of the input trace is in the support of the initial distribution of the program.

Then, the set of output traces of the label trace, $\sigma_{OUT}^\ell = \{\{\texttt{ego}:\langle\texttt{FL,FL,LC,FL}\rangle,$ $\texttt{otherCar}:\langle \texttt{St,St,}$$\texttt{St,St}\rangle\}, \{\texttt{ego}:\langle\texttt{LC,LC,LC,LC}\rangle,$ $\texttt{otherCar}:\langle \texttt{St,St,}$\\$\texttt{St,St}\rangle\}\}$ matches the program because both \texttt{ego} and \texttt{otherCar} have at least one shared output trace with the program's. In contrast, the output trace $\sigma_{OUT}^\ell = \{\{\texttt{ego}:\langle\texttt{TL,FL,LC,TL}\rangle,$ $ \texttt{otherCar}:\langle \texttt{St,St,St,St}\rangle\}, \{\texttt{ego}:\langle\texttt{LC,LC,LC,LC}\rangle,\\ \texttt{otherCar}:\langle \texttt{St,St,}$ $ \texttt{St,St}\rangle\}\}$ where \texttt{TL} represents turn left, does not match. This is because \texttt{ego} does not have any shared output trace. This means that, given the input trace, the observed output trace cannot be generated by the program. Thus, this is not a match. 

% In this case, the mismatch occurs at the last timestep where the intersection between the sets of the output traces of $\textit{l}$ and $P$ is empty, i.e. $\{\texttt{TL}\}\cap\{\texttt{FL}\}\neq\emptyset$. Note that in the second to last step, there exists an intersection between \texttt{ego}'s behavior from the two output traces, i.e. $\texttt{FL}\in\{\texttt{FL,TL}\}\cap\{\texttt{FL,LC}\}$. Thus, this is not a mismatch. Recall that $\sigma_{OUT}^l$ record predictions of feasible behaviors. This means that \texttt{ego}'s behavior can \textit{either} be \texttt{FL} \textit{or} \texttt{TL}. The fact that there exists an intersection at each timestep for each behavior means that there is a output trace of the program that is consistent with the observed behaviors of all objects in the label trace. Thus, it is a match. 

%% file: sections/problem.tex
We formalize the querying problem as a \emph{membership} problem, where a trace, defined over a predetermined set of labels, is queried against a scenario modeled by a \scenic program.
We begin by introducing some necessary notation. 

% In this section, we formalize the \textit{query} on whether a video sensor data (e.g. RGB, LiDAR) contains a scenario which is modeled as a \scenic{} program. In particular, we formalize the query over the label trace of the sensor data, not the sensor data itself. We assume that these labels are given. 

\indent\textbf{Notation.} 
We write $|X|$ for the cardinality of set $X$ and $X \dot\cup Y$ for the disjoint union.
If $V$ is a set of variables that are defined over domain $D$, we define a valuation of $V$ as a function $\nu\colon V \rightarrow D$, and write the set of valuations of $V$ as $D^V$. $Dist(D)$ is the set of distributions over $D$, $D^*$ is the set of tuples whose elements are in $D$, and $\mathcal{P}(D)$ is the power set of $D$.
For a sequence $\sigma = \alpha_0,\dots , \alpha_n$, we define $\sigma(j) = \alpha_j$ for $\forall j\leq n$. Finally, we call a sequence $\sigma[i, j] = \alpha_i,\dots,\alpha_j$ for $0\leq i\leq j \leq n$ a \emph{window} of $\sigma$.

% \subsection{Definition: A Hierarchical Behavior in \scenic{}}\label{definition:behavior}
% When modeling a scenario in \scenic, a stochastic, reactive behavior of an agent can be specified. We abstract away the stochasticity with non-determinism and formally represent the behavior as an interrupt-driven, hierarchical, non-deterministic finite state machine which consists of a tuple, \textit{IHFSM} = $(I, O, S, S_0,$ $ l, L, G, \tau)$. $I$ is a set of input variables. $O$ is a set of output variables. $S \subseteq (I \rightarrow O)$ is a finite set of states mapping inputs to outputs. $S_0$ is a set of initial states. A state $s\in S$ can be hierarchical, where $s$ is defined as another tuple, $IHFSM'$. $l: S\rightarrow L$ is a labeling function that maps states to a set of labels, $\sigma$. $G$ defines a set of guards, i.e. predicates with different priorities, over the states. A guard, $g\in G$, is a tuple of a predicate for transition and a non-negative number, $n\in \{\bot\}\cup\mathbb{N}$, where $\bot$ represents that no priority is assigned and a higher number has a higher priority. $\tau \subseteq S\times G\times S$ is the transition relation over states.

% $\gamma: S\times I\rightarrow O$ maps each state to a subset of the outputs. 

\indent\textbf{\scenic Programs.} For purposes of this paper, we formally define a \scenic program as a tuple $P = (\textit{Obj}, I,$ $ O, \textit{Init}, B,\Gamma)$. $\textit{Obj} = \{obj_1, \dots, obj_n\}$ is a finite set of objects. $I$ and $O$ are disjoint finite sets of input and output variables, defined over a domain $D$; the input variables $I$ encode the state of the world modeled by the \scenic{} program (comprising semantic features such as positions of objects), while the output variables $O= O_1 \dot\cup \dots \dot\cup O_n$ represent a set of names of {\it primitive behaviors} for each object (see Sec.~\ref{sec:overview}). $\textit{Init}\in \textit{Dist}(D^I)$ is an initial distribution over the valuations of input variables. Finally, $B$ is a set of behaviors and $\Gamma\colon\mathit{Obj}\rightarrow B\cup\{\bot\}$ maps each object to its behavior (or $\bot$ if it has none).

\indent\textbf{Behaviors.}
Each object $o = o_i\in \textit{Obj}$ has an associated (general) {\it behavior} $b_o\in B$ which is a function $b_o\colon {(D^I)}^* \rightarrow \mathcal{P}(D^{O_i})$ defining, given an input sequence, the set of possible primitive behaviors for that object at the current time step. 
%
%We define a (general) {\it behavior} $b_o\in B$ of an object $o = o_i\in \textit{Obj} $ as a function $b_o\colon {(D^I)}^* \rightarrow \mathcal{P}(D^{O_i})$ defining the set of possible primitive behaviors for a given sequence of inputs. 
We use a set rather than a distribution of outputs as we are only concerned with membership, so we may abstract the randomness of \scenic{} behaviors into nondeterminism.
Let a \emph{trace} be a sequence $\sigma= \langle( {i}_0,  {o}_0), \dots , ( {i}_m, {o}_m)\rangle$ for some $m\in \mathbb{N}$, where $ {i}_j$ and $ {o}_j$ are valuations of the input and output variables at time step $j$. 
A behavior $b_o$ induces a set of traces $T_o = \{\langle( {i}_0,  {o}_0), \dots , ( {i}_m, {o}_m) \rangle \mid \forall j\leq m.~i_j \in D^I, ~o_j\in b_o({i_0}, \dots,  {i_j})\}$, which we call the set of traces of $b_o$. 
For a trace $\sigma$, we write $\sigma_\textit{in}$ for the projection of $\sigma$ to its sequence of input valuations, i.e., $\sigma_\textit{in} =  \langle{i}_0, \dots,  {i}_m\rangle$. 
Similarly, we write $\sigma_\textit{out}$ for the sequence of output valuations, i.e., $\sigma_\textit{out} =  \langle{o}_0, \dots,  {o}_m\rangle$. \\
\indent\textbf{The Semantics of a Program.}
Given a \scenic{} program $P$ as above, we define the behavior of $P$ as the combined synchronous behaviors of its objects, i.e., the function $b_P\colon {(D^I)}^* \rightarrow \mathcal{P}({D^{O_1}) \times \dots \times \mathcal{P}(D^{O_n}})$, mapping $\sigma_\textit{in}\in {(D^I)}^*$ to $b_1(\sigma_\textit{in}) \times \dots \times b_n(\sigma_\textit{in})$. Given the input trace $\sigma_{in}=\langle i_0,\dots,i_m\rangle$, we then define the set of traces of $P$ as $T_{P} = \{\langle( {i}_0,  {o}_0), \dots , ( {i}_m, {o}_m) \rangle \mid i_0 \in \textit{Supp(Init)} \text{ and } \forall j\leq m.~ i_j \in D^I ,~o_j \in b_1( i_0,\dots,i_j) \times \dots \times b_n( i_0,\dots,i_j)\}$, where \textit{Supp(.)} defines the support of a distribution. We say that a trace $\sigma$ \emph{matches} a \scenic program $P$ if $\sigma \in T_{P}$. We further say that a set of traces $T$ matches the program $P$ if $T\cap T_{P} \neq \emptyset$. Recall in Sec.~\ref{sec:overview} that a label trace, which we formalize below, can define a set of \textit{feasible} observation traces. Thus, as long as there exists a feasible observation trace that can also be generated by the program, it is a match. Lastly, given $\sigma' \in (D^{I})^*$,  we define $T_P^{\sigma'}=\{ \sigma \in T_P \mid \sigma_\textit{in} = \sigma' \}$ which represent the set of program traces generated with the input trace $\sigma'$.\\
\indent\textbf{Label Trace.} In our problem formulation, we query the \textit{labels} of time series sensor data, not the data itself. For brevity, we refer to a sequence of labels of each frame of sensor data (e.g. RGB image, LiDAR 3D point cloud) as a \textit{label trace}. Formally, a label trace is defined as a tuple, $\ell = (Obj, I, O, \sigma_\text{in}, \Sigma_\textit{out})$. Similar to the definition of a \scenic  program, $Obj$ is a set of objects, $I$ and $O$ are sets of input and output variables, respectively, both of which are semantic variables of objects and $I\cap O=\emptyset$. The label trace also consists of an input trace $\sigma_{in} \in (D^I)^*$ and a \textit{set} of output traces $\Sigma_\textit{out} \subseteq (D^O)^*$.
%\footnote{Recall that in the Overview (Sec.~\ref{alg:overview}), we explained that this set of output traces derive from the uncertainties in predicting the primitive behaviors of objects using a behavior prediction model.}
Lastly, we define the set of traces induced by $\ell$ by $T_\ell = \{ \langle(i_0,o_0),\dots, (i_m,o_m) \rangle\mid \forall j\leq m.~\sigma_{in}(j) = i_j \wedge ~\exists \sigma_{out} \in \Sigma_{out}.~ \forall j\leq m.~ \sigma_{out}(j) = o_j\}$.\\
\indent\textbf{Problem Statement.}
Let a \scenic{} program, $P = (\mathit{Obj^P}, I^P, O^P, Init, B, \Gamma)$, and a label trace, $\ell=(\mathit{Obj^\ell}, I^\ell, O^\ell, \sigma_{in}, \Sigma_{out})$, be given. It is possible that the label trace can contain additional objects or observations compared to $P$. For example, a program modeling a left-turn scenario with two cars should match a trace containing two such cars, even if there is also an unrelated pedestrian in the trace and even if the trace continues after the left turn is complete. We allow the label trace to contain additional information, but do not allow the program to contain additional information. We formalize this by using the following notion of the object correspondence. \\
\indent To formalize the notion of a match between $P$ and $l$, we need to check whether the objects of $l$ behave as specified in $P$. This requires a mapping between objects in $P$ and $l$. Thus, we define an \textit{object correspondence} as an injective function $C\colon \mathit{Obj^P}\rightarrow \mathit{Obj^\ell}$, mapping the objects in the program, $Obj^P$, to those in the label trace, $\textit{Obj}^\ell$ (later in our methodology we show how such a mapping can be computed).
Given a trace $\sigma$, let $C^{-1}(\sigma)$ (slightly abusing notation) denote the trace obtained by mapping each object in $\sigma$ to its corresponding object in the program according to $C$, if there is one, and dropping it otherwise. We assume $P$ and $\ell$ share the same input and output variables (by restricting the sets to the common variables). \\
\indent\textbf{Problem:} 
For a \scenic{} program $P$, 
a label trace $\ell$,
and an integer $m \in\mathbb{N}$, check whether there is a correspondence $C$,  trace $\sigma \in T_\ell$, and length-$m$ window $\sigma'$ of $C^{-1}(\sigma)$ such that $\sigma' \in T_{P}^{\sigma'_{\textit{in}}}$.

%% file: sections/methodology.tex
Given a \scenic{} program and a label trace, our key idea is to translate the program to a synchronous composition of hierarchical finite state machines (HFSMs) (see~\cite{leeseshia-16}), where each HFSM represents a behavior of an object. The query problem then reduces to checking if the HFSMs accept some trace consistent with the label trace \cite{scenic_querying}. We solve this problem by extending the classical non-deterministic finite automata simulation algorithm to our symbolic HFSMs.
% Then, we determine if the synchronous HFSMs, i.e. the program, \textit{matches} the label trace. In this section, we give a formal representation of the behaviors of objects in a \scenic{} programs using input-output hierarchical finite state machines, and solve Problem P1 as one of checking whether the label trace mimics some of the input-output behaviors induced by the composition of machines. 
\begin{table}[t]
    \centering
    \begin{tabular}{|c|c|}
    \hline
        Supported Fragment Type& Syntax\\
        \hline
        Distribution & Uniform, Range, Normal, TruncatedNormal\\
        \hline
        Statements & require boolean, do behavior, do behavior until, try / interrupt\\
        \hline
        Position Specifier & at, in, on, offset by, beyond by, visible from, ahead of, \\
         & behind by, following for \\
        \hline
        Orientation Specifier & facing orientation, facing toward / away from, apparently facing\\
        \hline
        Scalar Operators & relative heading of, apparent heading of, distance to, angle to\\
        \hline 
        Boolean Operators & can see, in\\
        \hline
        Orientation Operators & deg, relative to\\
        \hline
        Vector Operators & offset by, offset along by \\
        \hline
        Region Operators & visible, not visible, visible from, not visible from\\
        \hline
    \end{tabular}
    \caption{Supported \scenic{} Fragment for Querying Label Traces}
    \label{table:scenic_fragment}
    \vspace{-0.5 cm}
\end{table}

\subsection{Supported \scenic{} Fragment}\label{sec:scenic_fragment}
We first specify the fragment of \scenic{}~\cite{scenic-journal} syntax that is supported in our methodology as it affects the definition of the HFSM. We support all of \scenic{}'s operators for scalar, boolean, orientation, vector, and region, as well as specifiers for position and orientation, but we restrict the types of statements as shown in Table~\ref{table:scenic_fragment}. We visualize our fragment related to modeling behaviors in Fig.~\ref{fig:hfsm}. This fragment allows one to flexibly specify any sequential or interrupt-driven behaviors, which can flexibly model diverse interactions among objects. On the other hand, our supported fragment permits variable assignments. This restriction simplifies behaviors so that they become memoryless, meaning a behavior maps an input (not a history of inputs) to a set of outputs. 

\subsection{Input-output hierarchical finite state machines}
Each behavior defined in the \scenic{} program is abstracted into a hierarchical finite state machine (HFSM)~\cite{hfsm-Yannakakis}. Formally, an HFSM is a tuple, $M= (I, O, S, S_0, \mu, l, G, \tau)$. The sets, $I$ and $O$, are the sets of input and output variables, respectively. $S$ is a finite set of states and $S_0\subseteq S$ is a set of initial states. The function $\mu\colon S\rightarrow \mathcal{M}\cup\{\bot\}$ is a mapping from states to HFSMs where $\mathcal{M}$ represents the set of all HFSMs, and $\bot$ means that a state is non-hierarchical. This $\mu$ defines the refinements of each state of the HFSM $M$. The labeling function, $l\colon S\rightarrow V^O$, maps states to concrete output values. $G$ is a set of guards, where a guard is a Boolean predicate defined over $I$. Finally, $\tau \subseteq S\times G\times S$ is the transition relation over states. \\
\indent Returning to our running example in Sec.~\ref{sec:overview}, recall that a behavior is hierarchically defined with a pre-defined set of discrete primitive behaviors in \scenic{}, e.g. follow lane, lane change, etc. The output variables, $O$, of the HFSM specify each object's feasible output behaviors. The input variables, $I$, specify the remaining observations of the world, such as positions, of all objects. As we explained in Sec.~\ref{sec:scenic_fragment}, our behaviors are memoryless. This means that guards are not dependent on the past input. For example, in the \scenic{} program in Figure~\ref{fig:scenic-program}, the guard of the condition in line 4 only depends on the current input. The guard predicates of HFSMs are encoded as a satisfiability modulo theories (SMT)~\cite{barrett-smtbookch09} formula with a non-linear real arithmetic theory. We provide context for the rest of the HFSM constructs in the following section. \\
\indent Note that, in Figure~\ref{fig:scenic-program}, \texttt{Range(1,15)} in line 4 is not a semantic feature variable in $I$. It is a variable instantiated by the program and its value is not provided for in the label trace. We refer to such variable as an \textit{unobserved} variable. In such a case, we encode the domain of such variable to the SMT formula of the guard, such that it evaluates to true if there exists a value in the domain that satisfies the formula. Assuming that the positions of \texttt{ego} and \texttt{otherCar} are observed at the current timestep, their distance on the left side of the inequality can be evaluated. If their distance is in between 1 and 15 meters, then the guard can be evaluated to be true; otherwise, false. Likewise, the guard condition for \texttt{FollowLane} is the \textit{negation} of the condition in line 4, which also can be evaluated to true if their distance is in between 1 and 15 meters. Thus, in such a case, non-deterministic transitions are executed such that \texttt{ego} can be in either \texttt{FollowLane} or \texttt{LaneChange} states at the current timestep. Thus, note that each object's behavior can have more than one feasible primitive behavior output due to the non-determinism.
\indent The semantics of the HFSMs for the supported fragment follow the semantics of a \scenic{} program as defined in the problem formulation. The overall behavior of the program is defined as the combined synchronous behaviors of all its objects. At the initial timestep, each HFSM initializes to its initial states. Given an observed input at each timestep, each HFSM executes to transition its states. Starting from the topmost current states, the HFSM iteratively evaluates the guards of the current states with the input and non-deterministically transitions if they evaluate to true. Then, the HFSM recursively traverses down to its current child states repeating the same procedure until a non-hierarchical state, i.e. a state with no child, is reached \cite{scenic_querying}. For brevity, we define a base state to be a state with no child state. Prior to reaching any base states, if a hierarchical state reaches its terminate state (as shown in Figure~\ref{fig:hfsm}) then its parent state transitions and then resumes the recursive procedure. Once the base states are reached, the HFSM outputs using the label function for the base state. If the topmost hierarchical state reaches a terminate state, then the whole HFSM terminates. In our running example, the base states are the discrete primitive behaviors, and the labeling functions for the base states return the string name of the base states, e.g. ``\texttt{FollowLane}.'' 
 
% From now on, we will use the term behavior and HFSM, interchangeably.

% A \textit{run} of $M$ for an input trace, $\sigma_{in}=\langle i_0, ...,i_T \rangle$, is a sequence of states and guards, $\rho= \langle s_0, g_1, s_1, ..., g_T, s_T \rangle\in S_0\times (G\times S)^*$, generated by the HFSM in the following way. The guards can have Boolean values: \textit{true} or \textit{false}. If the guard evaluates to true, then a state transition occurs. If it is false, the state would not transition. In $\rho$, it holds that for $\forall t \in \{0, \dots, T\}$, $g_t(i_t)= \text{true}$, and $s_i, g_{i+1}, s_{i+1} \in \tau$ for $\forall i\in\{0,...,T-1\}$. We further, define an output trace of $\rho$ as a sequence of labels of its states, $\langle l(s_0), ...,l(s_T) \rangle$, where $l$ is a labeling function of the $M$. Note that because the HFSM is non-deterministic, it may generate a set of multiple runs for $\sigma_{\textit{in}}$. 

\subsection{Syntax-Directed Translation to HFSMs}\label{sec:scenic_hfsm_translation}
\scenic{} provides syntax to hierarchically model behaviors using a set of primitive behaviors as shown in Figure~\ref{fig:scenic-program}. We translate the behaviors in the \scenic{} program to HFSMs, maintaining the hierarchical structure. Hence, we refer to it as a \textit{syntax-directed} translation to HFSMs. Given a \scenic{} program, we convert it into an abstract syntax tree and then translate it to a set of HFSMs.
\begin{figure}[tb]
    \centering
    \includegraphics[width=0.7\linewidth]{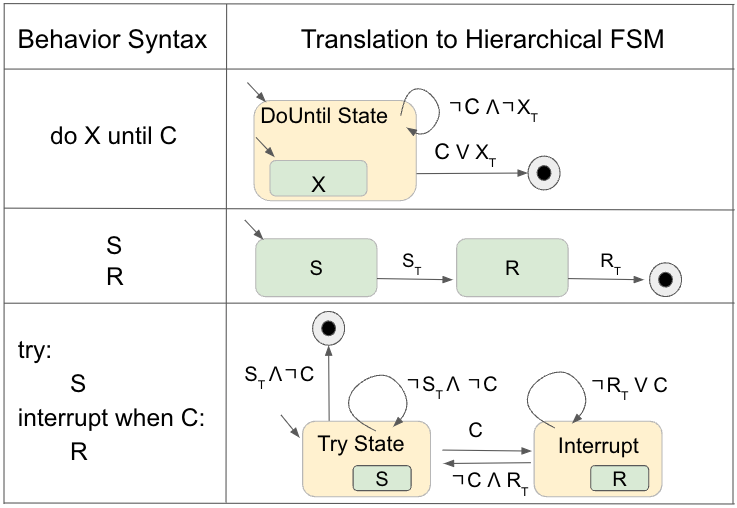}
    \caption{Translation of \scenic{} syntax fragment to hierarchical FSM. The green boxes abstractly represent finite state machines, while the yellow boxes are states. The state with the black dot represents a termination state. The variables with $T$ subscript are Boolean predicates which indicates whether the corresponding machine terminated.}
    \label{fig:hfsm}
\end{figure}

\indent Figure~\ref{fig:hfsm} shows the translation each behavior in a \scenic{} program to a HFSM. On the left column, a construct of our supported \scenic{} fragment for modeling behaviors are listed. On the right column, the corresponding HFSM translation is shown, using the traditional statechart~\cite{state-chart} convention to visualize HFSMs. Each box represents a state named after the \scenic{} syntax it represents. If a box contains another box, this represents that the box containing the other is the parent. The black double circle represents a termination state. The symbols on the arrows, i.e. transitions, represent the guards.
%The logical symbols, $\neg$, $\lor$, and $\land$, mean negation, disjunction, and conjunction of Boolean predicates, respectively.  

The first row of Figure~\ref{fig:hfsm} shows the translation of \texttt{do X until C} syntax which is used to invoke behavior \texttt{X} until a condition \texttt{C} is satisfied. On the right, the \texttt{DoUntil} state has a child called \texttt{X}, which recursively executes upon the execution of its parent state. A Boolean predicate, $\texttt{X}_\texttt{T}$, indicates whether the behavior \texttt{X} is completed. The black circle represents a termination state. A \texttt{do} statement, as shown in Figure~\ref{fig:scenic-program}, is equivalent to case where the condition \texttt{C} is assigned false. In the second row, a sequential statement is shown where \texttt{S} and \texttt{R} can be any statement shown in Figure~\ref{fig:hfsm}. On the right, \texttt{S} and \texttt{R} represent HFSM that represents the corresponding statements. Finally, the HFSM translation of the \texttt{try/interrupt} statement is shown on the third row, left. Again, \texttt{S} and \texttt{R} are statements of any type in the figure. The \texttt{Try} and \texttt{Interrupt} states contain the HFSMs of the corresponding \texttt{S} and \texttt{R} statements. In our methodology, primitive behaviors are compiled to base states. For example, in Fig.~\ref{fig:scenic-program}, \texttt{FollowLane}, \texttt{LaneChange}, and \texttt{Stationary} are base states.
The behaviors, \texttt{X}, \texttt{S}, and \texttt{R}, in Figure~\ref{fig:hfsm} can be either a hierarchical state, i.e. HFSM, or a base state.

\subsection{Query Algorithm}
Our top level algorithm is given in Algorithm~\ref{alg:overview}. As explained in Sec.~\ref{sec:scenic_hfsm_translation}, the program $P$ is converted into an abstract syntax tree (AST) by \texttt{Parse} function (line 1), and the AST is translated to HFSMs by \texttt{Translate} function (line 2). Because the \emph{object correspondence} injectively mapping objects in the \scenic{} program to those in the trace is unknown \emph{a priori}, we begin by searching for this mapping.

\indent\textbf{Correspondence Search.}
The algorithm analyzes the abstract syntax tree (AST) of the program to identify all the objects and their types (e.g. pedestrian, car). Likewise, from the label trace, it extracts all the observed objects, their types, and the duration for which they have been observed.
% \footnote{Note that not all objects may be observed for all timesteps as some of them may move out of the visible area. For example, imagine a self-driving car moving through San Francisco. The observed pedestrians and vehicles would disappear from the car's visible area as the car navigates through the city}. 
Then, for each \scenic{} object, the algorithm identifies a set of objects in the label trace that share the same object type and are observed for at least the provided minimum time duration $m$. This way, the algorithm aims to prune out infeasible correspondences, thereby reducing the number of combinatorial searches. These procedures are executed by the function in line 3, which encodes these constraints as a satisfiability modulo theory (SMT) formula, $\phi$, with the linear integer arithmetic logic. Returning to our running example, the SMT formula may encode that $\texttt{ego}, \texttt{otherCar}\in\{\textit{car1, car5, car10, car21}\}$ and that $\texttt{ego}\neq \texttt{otherCar}$, where \textit{car\#} refers to a unique integer id\# in the label trace.

\begin{algorithm}[tb]
\caption{Determine if a \scenic{} program matches a label trace}
\label{alg:overview}
\textbf{Input}: \scenic{} program $P$, a label trace $l$, and a minimum time duration $m$\\
\textbf{Output}: Does $l$ match $P$? (True / False) 
\begin{algorithmic}[1]
\STATE $\textit{AST} \gets \textit{Parse}(P)$ \COMMENT{get abstract syntax tree (AST)}
\STATE $M\gets\textit{Translate}(AST)$ \COMMENT{$M=\{\text{key:object name, value:HFSM of object's behavior}\}$}
\STATE $\phi \gets \textit{InitializeCorrespondenceSMTConstraints}(l,AST,m)$ 
%\STATE $\textit{corr}\gets \textit{SMTSolver}(\phi)$// an object correspondence
\WHILE{$\phi$ is satisfiable, with solution \textit{corr}}
\IF[Algorithm 2]{$Query(AST,M,l, \textit{corr}, m)$}
\STATE \textbf{return} \textit{True}
\ENDIF
\STATE $\phi \gets \phi \wedge \phi'$ \COMMENT{$\phi'$ is a new constraint to block already checked object assignment}
%\STATE $\textit{corr} \gets \textit{SMTSolver}(\phi)$ 
\ENDWHILE 
\STATE \textbf{return} \textit{False}
\end{algorithmic}
\end{algorithm}

If the SMT solver returns \textit{unsat}, meaning there is no solution, then there is no feasible correspondence. Thus, the algorithm returns \textit{False}, i.e. not a match. If there exists a correspondence, then we use the correspondence to determine a match (line 5). If the label trace is not a match, then it is possible that there still may be another correspondence which may result in a match. Thus, as shown in line 7, the algorithm conjoins an additional constraint $\phi'$ which encodes that the current correspondence is not true to the SMT formula $\phi$, e.g. $\phi' = \neg(\texttt{ego}=\textit{car5}\land \texttt{otherCar}=\textit{car3})$. Then, the algorithm searches for another correspondence using the SMT solver until either a match is found (returning True) or until no more correspondence can be found (returning False).

\begin{algorithm}[tb]
\caption{Query Algorithm}
\label{alg:query}
\textbf{Input}: Abstract Syntax Tree of the \scenic{} program $AST$, Compiled HFSMs $M$, label trace $l$, object correspondence \textit{corr}, and time duration $m$ \\
\textbf{Output}: $l$ matches given $corr$
% Whether $M$ accepts a length-$m$ window of $l$ given $corr$ ???
\begin{algorithmic}[1]
\FOR{timestep \textit{i} from 0 to $len(l)-m$}
\IF{$\textit{InitialInputMatch}(AST, l, t)$ is False} 
\STATE \textbf{continue} // initial scene does not match; try next window
\ENDIF
\STATE $\text{currentBaseStates} \gets \text{initial base states of $M$}$
\STATE // dictionary whose key is obj name and value is current base states of the obj's HFSM
\FOR{timestep \textit{t} from $i$ to $i+m-1$}
\STATE currentBaseStates $\gets$ ValidStep(currentBaseStates, $M$, $l[t]$, $corr$)
\IF{currentBaseStates[obj] is empty for any obj} 
\STATE \textit{mismatch} $\gets$ \textit{True}
\STATE \textbf{break} // mismatch detected, break out of the inner for-loop 
\ENDIF
\ENDFOR
\IF{not \textit{mismatch}}
\STATE \textbf{return} \textit{True}\ENDIF
\ENDFOR

\STATE \textbf{return} \textit{False}
\end{algorithmic}
\end{algorithm}

\indent\textbf{Query Procedure} The pseudocode for determining a match is shown in Algorithm~\ref{alg:query}. Because the problem is to find a match for $m$ consecutive timesteps, we use a sliding window of length $m$ across the label trace (line 1 and 6). For the label trace to match the program, its initial input must be in the support of the program's initial distribution.
The \texttt{InitialInputMatch} function on line 2 checks this condition using the algorithm from our prior work~\cite{iccps}.
If the check fails, then the algorithm moves on to the next sliding window (line 3). Otherwise, the algorithm proceeds to compare the output traces of the label trace and the \scenic{} program. \\
\indent The line 6-9 of the algorithm checks if the HFSM can simulate the label trace for a sliding window of length $m$. At each timestep of the sliding window, the algorithm executes \texttt{ValidStep} function which returns the set of possible outputs from each HFSM that are consistent with observed outputs from the label trace, i.e. the returned set is the intersection of the set of outputs from each \scenic{} object's HFSM and the set of outputs from the corresponding object in the label trace. We will describe \texttt{ValidStep} procedure in the next section and move on to explain the rest of Algorithm~\ref{alg:query}. For all timesteps of the sliding window of length $m$, if there exists a consistent output between the HFSMs and the output traces of the label trace, then the algorithm returns \textit{True}, i.e. a match; otherwise, the algorithm moves on to check the next sliding window. If all possible sliding windows are checked but none of them results in a match, then it outputs \textit{False}. Note that $m$ is a parameter which needs to be carefully chosen by a user. If $m$ is too small, e.g. m = 1, then it may likely return many label traces which may match for a single timestep but does not match for the most part of the traces.\\
\indent\textbf{ValidStep Procedure}
The \texttt{ValidStep} function steps, or transitions, the HFSMs such that their outputs are consistent with the observed outputs in the label trace. The function takes as an input argument a dictionary called $currentBaseStates$ whose key is an object name and its value is a set of feasible base states. The current base states refer to base states that are running at the given timestep. Recall that the output values of each HFSM are determined by the current base states (refer to the semantics of HFSMs in Sec.~\ref{sec:scenic_hfsm_translation}). In line 4, prior to invoking the \texttt{ValidStep}, all the HFSMs are initialized such that their current states are set to their initial states. Then, the current base states of each HFSM is computed by simply traversing down from the topmost current hierarchical states to the base states, without evaluating guards or transitioning states. Thus, the current base states of the HFSMs at the initial timestep of the current sliding window is computed. \\
\indent The \texttt{ValidStep} procedures are as follows. First, given the current base states of each HFSM, the function reconstructs the current states of the HFSMs by recursively traversing up each HFSM from its current base states. Then, starting from the topmost current states of each HFSM, the function recursively computes the following down their current child states. It encodes as SMT formula, with non-linear real arithmetic theory, each current state's guard conditions using the given input values at the current timestep, $l[t]$, uses a SMT solver to evaluate the guard, and transitions the states if any guard evaluates to true.\\
\indent Finally, once it transitions the HFSMs, the \texttt{ValidStep} function compares the set of outputs of each object's HFSM to the \textit{observed} outputs of the corresponding object in the label trace at the current timestep. Then, for each HFSM, the function \textit{prunes out}, i.e. deletes, its current base states whose outputs are not in the set of observed outputs. Thus, the function returns the \textit{pruned} set of current base states of each HFSM, which are consistent with the observed outputs. The returned set for a HFSM can be an empty set, which means that the HFSM's outputs are not consistent with the observed outputs, thereby not a match (line 8-10).

\subsection{\textbf{Correctness of the Query Algorithm}}
\begin{theorem}
    Given a \scenic{} program, a label trace, and an integer $m\in\mathbb{N}$, our algorithm outputs \textit{True} if and only if the label trace matches the program for a window of length $m$; otherwise, the algorithm outputs \textit{False}.
\end{theorem}

\noindent\textit{Proof Sketch.} Alg.~\ref{alg:overview} checks all possible object correspondences and windows of length $m$. For each correspondence, Alg.~\ref{alg:query}, in line 2, correctly checks whether the initial input is in the support of the initial distribution. With a synchronous composition of HFSMs as a formal representation of the provided program, the algorithm validates if the HFSMs matches the label trace matching its window by iteratively computing the set of HFSM states reachable after each prefix of the window, in line 6 of Alg.~\ref{alg:query}, as in the standard NFA simulation algorithm~\cite{nfa_sim}, which involves executing an NFA on an input string to determine if it reaches an accepting state.\footnote{Unlike Deterministic Finite Automata, NFAs can transition to multiple states at once due to non-deterministic transitions.} Our algorithm begins from the initial base states of each HFSM. Then, it synchronously updates the current base states by transitioning each HFSM to the next available states whose outputs are consistent with the outputs observed in the label trace at the given timestep. Thus, by induction over the length of the window, the algorithm results in computing the set of all possible output traces of the program that are consistent with those in the label trace. If the computed set is not empty, this means that there exists an observed output trace that is a member of the program's feasible set of output traces. Refer to Appendix~\ref{appendix:proof} for a detailed proof. 

\subsection{An Example Query}
Suppose an \scenic{} program in Fig.~\ref{fig:scenic-program}, a label trace whose contents are visualized as a Table~\ref{table:label_trace}, and a match duration timestep of $m=5$ are given as inputs for the query. The problem is to determine if the label trace contains the scenario in the \scenic{} program for five consecutive timesteps. 

\begin{figure}[t]
    \centering   \includegraphics[width=\linewidth]{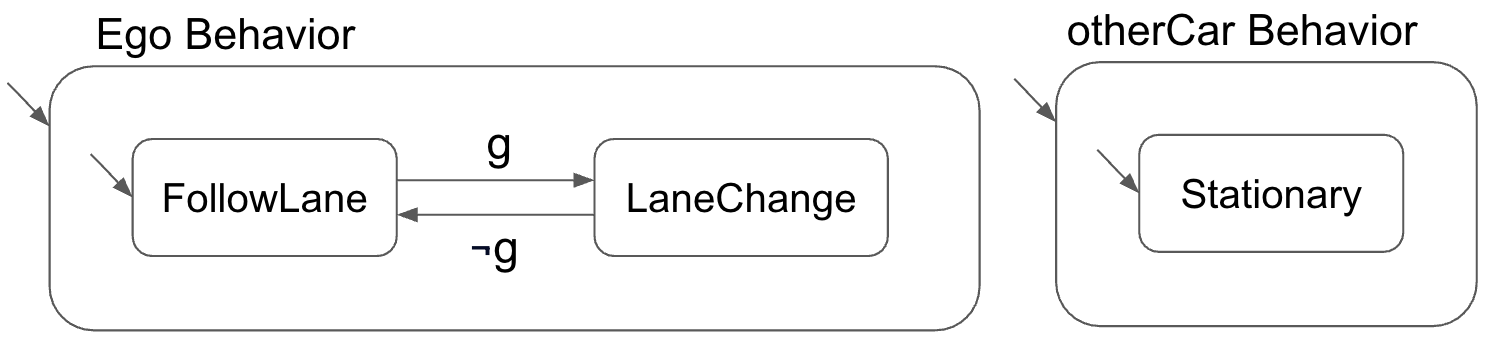}
    \caption{A hierarchical finite state machine representing the \scenic{} program in Fig.~\ref{fig:scenic-program}. The guard, $g$, represents the interrupt condition, i.e. (distance from ego to otherCar) < Range(1,15) meters.}
    \label{fig:example_fsm}
\end{figure}

\begin{table}
\centering
\begin{tabular}{| c |c | c | c | c | c| c|} 
 \hline
 TimeStep & Car1 Position & Car2 Position & Car1 Behavior & Car2 Behavior & Car1 Lane & Car2 Lane\\ 
 \hline\hline
 0 & (0, 0, 0) & (0, 0, 20) & \{Stationary\} & \{FollowLane\} & Lane1 & Lane1 \\ 
\hline
 1 & (0, 0, 0) & (0, 0, 14) & \{Stationary\} & \{FollowLane\} & Lane1 & Lane1  \\ 
\hline
 2 & (0, 0, 0) & (0, 0.3, 10) & \{Stationary\} & \{LaneChange\} & Lane1 & Lane1  \\ 
\hline
 3 & (0, 0, 0) & (0, 0.6, 6) & \{Stationary\} & \{LaneChange\} & Lane1 & Lane2  \\ 
\hline
 4 & (0, 0, 0) & (0, 1, 1) & \{Stationary\} & \{FollowLane\} & Lane1 & Lane2  \\ 
\hline
\end{tabular}
\caption{An example of a label trace whose contents are summarized in this table. Two vehicles are observed and recorded in this label trace.}
\label{table:label_trace}
\end{table}

\vspace{-1cm}
\begin{table}
\centering
\begin{tabular}{| c |c | c | c | c |c |} 
 \hline
timestep & distance & guard value & otherCar HFSM output & ego HFSM output & match?  \\ 
 \hline\hline
0 & 20 & False & \{Stationary\} & \{FollowLane\} & True  \\  
\hline
1 & 14 & True, False & \{Stationary\} & \{FollowLane, Lane Change\} & True  \\  
\hline
2 & 10 & True, False & \{Stationary\} & \{FollowLane, Lane Change\} & True  \\  
\hline
3 & 6.03 & True, False & \{Stationary\} & \{FollowLane, Lane Change\} & True  \\  
\hline
4 & 1.41 & True, False & \{Stationary\} & \{FollowLane, Lane Change\} & True  \\  
\hline
\end{tabular}
\caption{The computed value of guard condition, $g$, in Fig.~\ref{fig:example_fsm} based on the observed vehicle positions in the label trace (Table~\ref{table:label_trace}), and the computed HFSM outputs at each timestep.}
\end{table}

\vspace{-0.5cm}

The algorithm first parses and translates \scenic{} program to HFSMs(Alg.~\ref{alg:overview}, line 1-2) shown in Fig.~\ref{fig:example_fsm}, each modeling an agent's behavior. The label trace and the scenario program both contain two vehicles, respectively. There are two possible object correspondences (Alg.~\ref{alg:overview}, line 3): (i) \{ego = car1, otherCar = car2\} and (ii) \{ego = car2, otherCar = car1\}. Suppose the first correspondence (i) is selected. Since $m=5$ and the length of the label trace is also five, there is only a single sliding window to check match for in Alg.~\ref{alg:query}. In this case, querying with this correspondence results in a mismatch because \texttt{ego} cannot output stationary behavior, while \texttt{car1} in the label trace only exhibits stationary behavior. This results in the Alg.~\ref{alg:query} returning False. This correspondence is blocked (Alg.~\ref{alg:overview}, line 7), and the only remaining correspondence (ii) above is computed. Assuming that the initial scene is correct,\footnote{This initial scene check would require orientation values in the label trace, which we intentionally overlook to simplify the example.} the observed behaviors of \texttt{ego} and \texttt{otherCar} are subsets of the possible output by the corresponding agent's HFSM, respectively, for five consecutive steps. Thus, the algorithm outputs True. Note that the value of the guard $g$ can be both True and False from timestep 1 to 4 depending on the sampled value of \texttt{Range(1,15)} which is a uniform random distribution from a continuous interval of [1,15]. In this case, since both $g$ and $\neg g$ can be evaluated to True, the possible outputs from the HFSM of \texttt{otherCar} is both FollowLane and LaneChange.

% \noindent \text{Assigning Car A to Ego, CarB to OtherCar}
% \begin{table}
% \centering
% \begin{tabular}{|c | c | c | c |c |} 
%  \hline
% distance & transition cond & ego possible & other possible & match?  \\ 
%  \hline\hline
% 20 & False & (FollowLane) & (Stationary) & False  \\  
% \hline
% \end{tabular}
% \end{table}

%% file: sections/experiment.tex
% \sanjit{
% Suggest the following sub-sections for evaluation:\\
% 1. Comparison of Accuracy  with LLMs (current Table 1)\\
% 2. Efficiency of the Querying algorithm (how it scales with length of trace, complexity of scenario - num objects/agents)\\
% 3. Sim-to-Real Validation use case\\
% 4. Data analysis/understanding use case -- identifying problems with labeled trace data, etc.\\
% }

%Recall in the introduction section, we motivated the query problem with the application of validating failure scenarios identified in simulation. 
We evaluate our query algorithm for validating failure scenarios identified in simulation.
Once we can retrieve a set of video sensor data that match the failure scenarios, then we can test the components of autonomous systems, such as perception, behavior prediction, and planner, for sim-to-real validation. This application presupposes two hypotheses: (1) our algorithm is accurate, and (2) it can scale to complex queries. To validate these hypotheses, we conduct two different experiments. The first experiment focuses on evaluating the accuracy of the algorithm. The second experiment assesses the scalability of the query algorithm with respect to the size of the \scenic{} program and the duration of the videos that are being queried. In both experiments, we use the cvc5 SMT solver~\cite{cvc5} with its default parameter settings to compute guard conditions of HFSMs based on the input trace of the label trace.

\subsection{Accuracy Experiment}
We compare the accuracy of our query algorithm to two state-of-the-art vision language models (VLMs), GPT-4o~\cite{gpt4o} and Claude-3.5~\cite{claude}, which are developed for visual query and answer (VQ\&A)~\cite{vqa}. These models can answer natural language questions about RGB videos. We exclude other potential baselines, such as existing video retrieval database systems, as they cannot handle the level of expressiveness of queries that \scenic{} programs model (refer to related work in Sec.~\ref{sec:intro_related}). 

\begin{figure}[tb]
    \centering
    \includegraphics[width=0.7\linewidth]{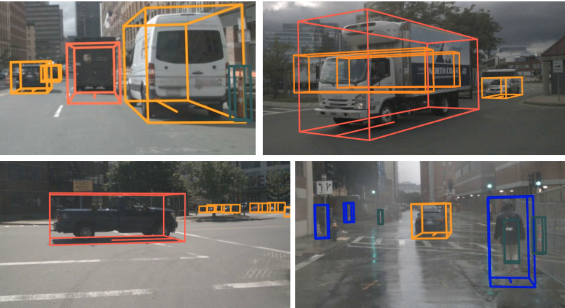}
    \caption{A snapshot from a matching video per scenario from Scenario 1 through 4 (clockwise from left top).}
    \label{fig:scenarios}
\end{figure}

\textbf{Scenarios.} Here we provide brief natural language descriptions of the scenarios. The images from matching videos for the four scenarios are shown in Figure~\ref{fig:scenarios}. 
\begin{enumerate}
    \item A car follows a lane then changes the lane to avoid a stationary car in front.
    \item A car is initially stationary waiting for an unprotected left turn. It waits for another car on the opposing lane to pass by and then makes the left turn.
    \item A car follows a lane and passes a pedestrian walking on the ego's lane. 
    \item A car yields to another car before making a right turn.
\end{enumerate}

\textbf{Data.} We use a subset of 400 RGB videos from nuScenes~\cite{nuscenes}. These data are collected in Boston while driving around the city in a vehicle with various sensors including RGB camera. nuScenes provide 20 second RGB video clips consisting of 40 image frames. Each image frame of the videos has a label containing information, such as positions, orientations, types (e.g. pedestrian, vehicle), and occupying lanes of observed objects. We post-process the RGB data and the label trace to compute the primitive behaviors, using a behavior prediction model~\cite{visiontrap} to classify the set of feasible primitive behaviors of observed objects in the videos (refer to label trace in Sec.~\ref{sec:overview}). We threshold the confidence scores of the model's classifications to record a set of feasible primitive behaviors of objects at each timestep. For each of the four scenarios, we prepare a set of five nuScenes labeled RGB videos to query. We manually confirm that all of the four sets contain matching videos for their corresponding scenarios. To prepare these videos, we watch through the 400 videos in the nuScenes dataset to identify matching videos and then randomly select the non-matching videos such that each set contains five videos. The reason for this limited number of videos per scenario is due to the manual process involved in checking the accuracy of the VLMs as explained below. 

\textbf{Query.} We manually encode the four scenarios as \scenic{} programs. Using our algorithm we query each of the four sets with its corresponding \scenic{} program and the label traces of its videos. To evaluate the VLMs, from each of the four sets, we input a RGB camera video and a natural language prompt which includes a description of a scenario as specified in the corresponding \scenic{} program and an instruction to (i) determine if the video contains the scenario (True/False) for at least half of the video duration, and (ii) if it contains the scenario, then provide the correspondence between the objects in the scenario description and the video by returning an image per object with an explanation on which object in the image belongs to the object in the scenario description. We ask for (ii) to prevent LLMs from randomly answering our question without understanding the video. We consider the LLMs answer to be accurate if they answer both questions (i) and (ii) correctly. For more details, refer to Appendix~\ref{appendix:accuracy-experiment}.

\begin{table}[tb]
\centering
\begin{tabular}{|c | c | c | c |} 
 \hline
 Scenario \# & Claude & GPT-4o & Our Algorithm \\ 
 \hline\hline
 1 & 0.4 & 0.2 & \textbf{1.0} \\ 
 2 & 0.2 & 0.6 & \textbf{0.6} \\
 3 & 0.6 & 0.8 & \textbf{1.0} \\
 4 & 0.6 & \textbf{0.8} & 0.6 \\
 \hline
 average accuracy (out of 1) & 0.45$\pm$ 0.19 & 0.60$\pm$ 0.28 & \textbf{0.80$\pm$ 0.23}\\ 
 \hline
 average runtime (sec) & 6.33$\pm$ 1.04 & 41.19$\pm$ 27.57 & \textbf{0.06$\pm$0.07} \\
 \hline
\end{tabular}
\caption{Comparison of video query accuracy between the state-of-the-art vision language models and our algorithm.}
\label{table:query_accuracy}
\vspace{-0.5cm}
\end{table}

\textbf{Results \& Analysis.} 
The experiment results are summarized in Table~\ref{table:query_accuracy}. In the table, the average accuracy of the VLMs and our algorithm are shown per scenario. The maximum accuracy is 1, meaning that the VLMs or the algorithm correctly query all five videos per scenario, whereas the lowest accuracy is zero meaning it fails to query all five videos. The result shows that, average, our algorithm is more accurate with query execution time that is orders of magnitude faster than the state-of-the-art VLMs. Our algorithm is $35\%$ more accurate than Claude and $20\%$ more than GPT-4o. Also, the algorithm takes approximately 0.06 seconds on average to query all 20 videos (= 4 scenarios x 5 videos), while Claude takes 6.33 seconds and GPT-4o takes 41.19 seconds. A key takeaway of this result is that, by using label information that are readily available in open-source datasets, we can query and retrieve label traces (and, thus, video sensor data) orders of magnitude faster with higher accuracy than the baselines. 

While the algorithm accurately queries for Scenario \#1 and \#3, we observe inaccuracies in the remaining scenarios. Analysis shows that these are due to errors in the labels pertaining to the predictions of the primitive behaviors of observed objects. For example, in Scenario \#2, we find that the primitive behaviors of a vehicle in the label trace unrealistically switches from left turn to right turn, while the vehicle actually executes an unprotected left turn in the RGB video. This illustrates that, although our algorithm is correct, its accuracy hinges on the accuracy of the labels. Similar label errors attribute to the inaccuracies of the algorithm in Scenario\# 4. 

\subsection{Scalability Experiment}
We evaluate the scalability of our algorithm with respect to (i) the duration of queried videos and (ii) the size of the query program. For (i), because videos in nuScenes dataset are clipped into 20 second videos, we generate synthetic label traces of varying lengths from 20 to 100 timesteps with a \scenic{} program (Figure~\ref{fig:scenic-program}) and the CARLA simulator~\cite{carla}. For (ii), we increase the size of the query program by replicating the objects and their behaviors in Figure~\ref{fig:scenic-program}, thereby increasing the number of objects from 2 to 8 objects (refer to Appendix~\ref{appendix:scalability}). Then, we generate synthetic label traces of 100 timesteps of duration with these programs and CARLA. For both experiments, the input parameter, $m$, to our algorithm is set to be the half of the queried label trace length, and we generate 10 label traces for each duration or number of agents. Note that these label traces match the corresponding programs by definition. 

\begin{figure}[tb]
    \centering
    \includegraphics[width=0.9\linewidth]{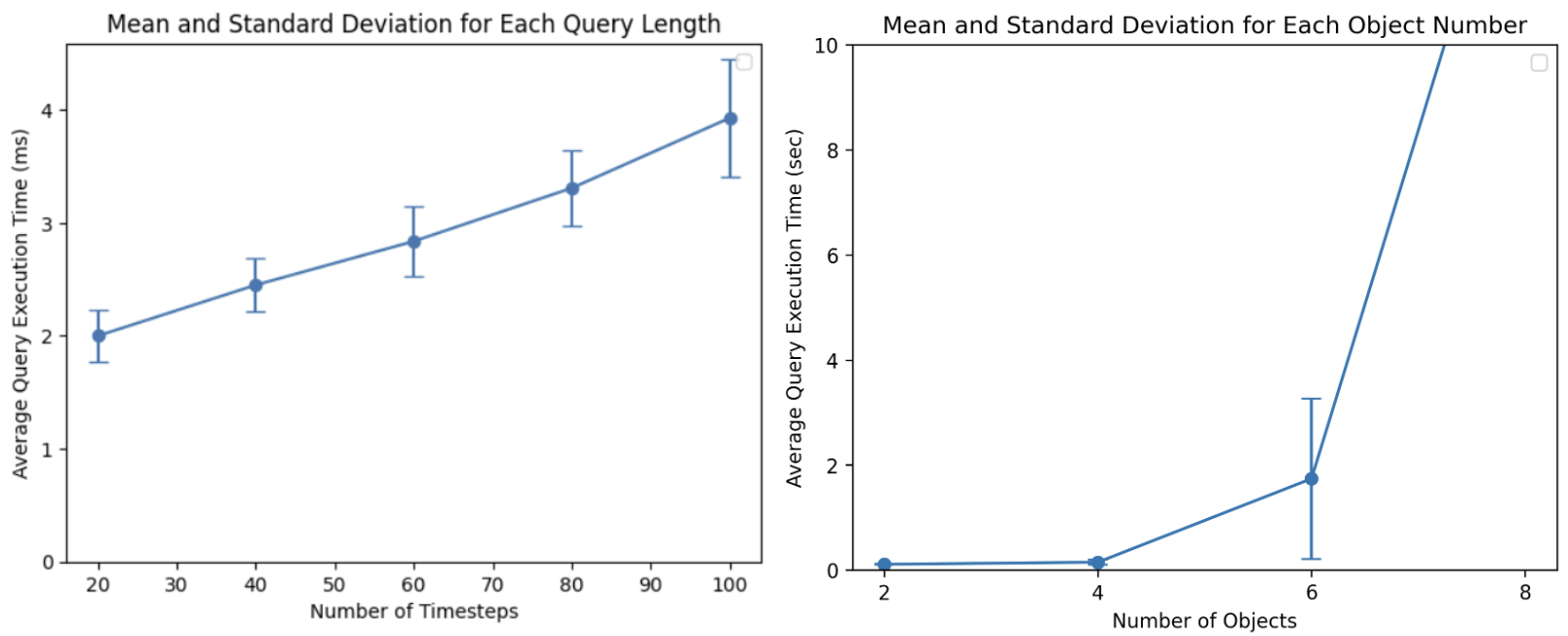}
    \caption{Scalability experiments of query runtime with respect to the increasing duration of label traces (left) and the number of objects in the program (right)}
    \label{fig:scalability-time}
    \vspace{-0.5cm}
\end{figure}

Figure~\ref{fig:scalability-time} summarizes the results. Each data point reported in the plots represents average query runtime over 10 label traces. The error bar over each data point represents the standard deviation. On the left of the figure, it shows that our query algorithm's runtime scales approximately linearly with respect to the duration of the queried label traces. Note that, in this experiment, there are only two object correspondences to check given that there are only two agents in the program. Thus, most of the query time is spent on either rejecting the wrong correspondence once or matching the correct one once over the increasing duration of time, resulting in the linear trend. In contrast, the right plot of Figure~\ref{fig:scalability-time} shows that our algorithm suffers from exponential increases in query time with respect to the number of objects in the program. For the query involving eight objects, the query times out at 10 seconds. This exponential trend is explained by the combinatorial correspondence search process (Alg.~\ref{alg:overview}). In case of the query involving eight objects, it needs to search over 8! = 40320 object correspondences in worst case.

%% file: sections/conclusion.tex
We present a novel query formulation to retrieve labeled time series sensor data, regardless of the data types (e.g. RGB, LiDAR, radar). Furthermore, we provided a query algorithm with a correctness guarantee to solve the problem. Our experiments show that the algorithm can query videos of labeled time series sensor data more accurately and orders of magnitude faster than the state-of-the-art vision language models, and can scale linearly to the duration of videos. However, our experiment also showed the limitations of the algorithm. In particular, the query runtime scales exponentially with respect to the number of objects. In the future, we aim to enhance the current combinatorial correspondence search process and also support a larger \scenic{} fragment to enable richer queries to better assist use cases such as sim-to-real validation.

\section*{Acknowledgments}

This work was supported in part by NSF POSE grant 2303564, Toyota and Nissan under the iCyPhy center, DARPA Contract HR00112490425 (TIAMAT), Berkeley Deep Drive, the Wallenberg AI, and Autonomous Systems and Software Program (WASP) funded by the Knut and Alice Wallenberg Foundation. The second author's work on this paper was done while he was a student at UC Berkeley. 

% In addition, we plan to conduct case studies of the algorithm pertaining to sim-to-real validation as well as assisting researchers to explore and understand the diversity of scenarios represented in a given big labeled dataset. 

%% file: sections/appendix.tex
% \section{Supported \scenic{} Fragment}\label{appendix:scenic_fragment}
% Our supported \scenic{} fragment is summarized in Table~\ref{table:scenic_fragment}.
% \begin{table}
%     \centering
%     \begin{tabular}{|c|c|}
%     \hline
%         Supported Fragment Type& Syntax\\
%         \hline
%         Distribution & Uniform, Range, Normal, TruncatedNormal\\
%         \hline
%         Statements & require boolean, do behavior, do behavior until, try / interrupt\\
%         \hline
%         Position Specifier & at, in, on, offset by, beyond by, visible from, ahead of, behind by, following for\\
%         \hline
%         Orientation Specifier & facing orientation, facing toward / away from, apparently facing\\
%         \hline
%         Scalar Operators & relative heading of, apparent heading of, distance to, angle to\\
%         \hline 
%         Boolean Operators & can see, in\\
%         \hline
%         Orientation Operators & deg, relative to\\
%         \hline
%         Vector Operators & offset by, offset along by \\
%         \hline
%         Region Operators & visible, not visible, visible from, not visible from\\
%         \hline
%     \end{tabular}
%     \caption{Supported \scenic{} Fragment for Querying Label Traces}
%     \label{table:scenic_fragment}
% \end{table}

\section{Proof of Theorem 1}\label{appendix:proof}
According to our problem statement, the label trace $l$ matches the program $P$ if there exists a window of the label trace of length $m\in\mathbb{N}$, whose (1) initial input is in the support of the initial distribution of the program, and (2) the window is a member of the set of program traces generated by the inputs of the window. 

The condition (1) is checked in Algorithm~\ref{alg:query} line 3-4 as the result of invoking the function, \texttt{Query}, in line 6 in Algorithm~\ref{alg:overview}. Thus, we move on to prove that algorithm correctly checks condition (2). 

For condition (2), we assume that our translation from the \scenic{} program to the HFSMs is accurate. This means that, given any input trace, the sets of all possible output traces from the program and those produced by the HFSMs are equivalent. Thus, under this assumption, checking condition (2) above is equivalent to checking that the translated HFSMs of the program can generate a length-$m$ subtrace of the label output trace. Recall that the \scenic{} fragment we support (as explained in Sec.~\ref{table:scenic_fragment}) prevents any variable assignments except for objects. Therefore, it is not allowed in our \scenic{} fragment to define and use a variable in any guard conditions such that the evaluation of the guards require access to the history of inputs. In short, the supported \scenic{} fragment computes the guards only based on the current inputs. For this reason, we can make the following inductive argument.

By induction over the length of the window $m$, the invariant of the induction step is that the \textit{current} states of the HFSMs are the reachable states which are consistent with the observed output of the label trace at each timestep. As a base case, at the initial timestep, $t_0$, the current states of the HFSMs are initialized to the initial base states (Alg.~\ref{alg:query} line 4). Then, for each HFSM, the \texttt{ValidStep} function traverses from the topmost initial hierarchical states down to the base states. As the function traverses to each current state, it evaluates all guards of the state based on the current input values and then transitions if any guards become true. Given that the guards are independent of the prior inputs, the function's evaluation of guards based on only the current input is correct. Thus, \texttt{ValidStep} correctly result in first identifying all the reachable base states at $t_0$. Then, the function applies the pruning of base states to ensure that the reachable base states are consistent with the observed output for each object's HFSM.  By induction, if we iteratively apply \texttt{ValidStep} from $i$ to $i+m-1$ and the returned current base states from the function is not empty for all HFSMs, then there exists an execution of HFSMs which are consistent with the output trace from time $i$ to $i+m-1$. Thus, by definition, the label trace is a match. Otherwise, it is not.

The algorithm enumerates the match procedure over all feasible correspondences and windows of length $m$ of the label trace. If there exists a correspondence and a window that result in a match, then the algorithm terminates and correctly outputs true; otherwise, it enumerates all possible combinations of choices and, thus, outputs false. 

\section{On Accuracy Experiment}\label{appendix:accuracy-experiment}
\subsection{\scenic{} Programs} 
\begin{figure}
    \centering
    \includegraphics[width=0.8\linewidth]{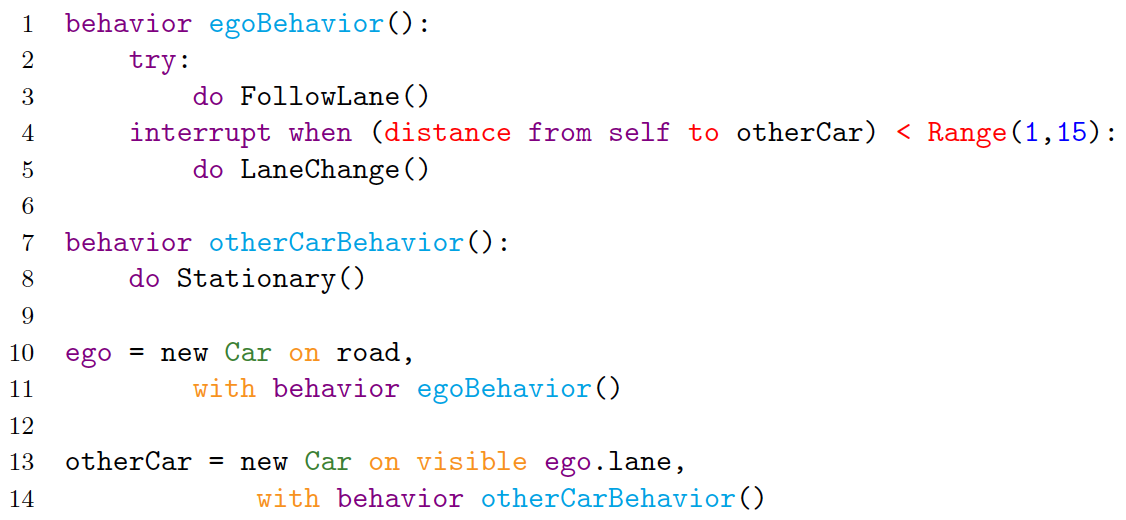}
    \caption{Accuracy Experiment (Scenario \#1)}
    \label{fig:scenario1}
\end{figure}

\begin{figure}
    \centering
    \includegraphics[width=0.8\linewidth]{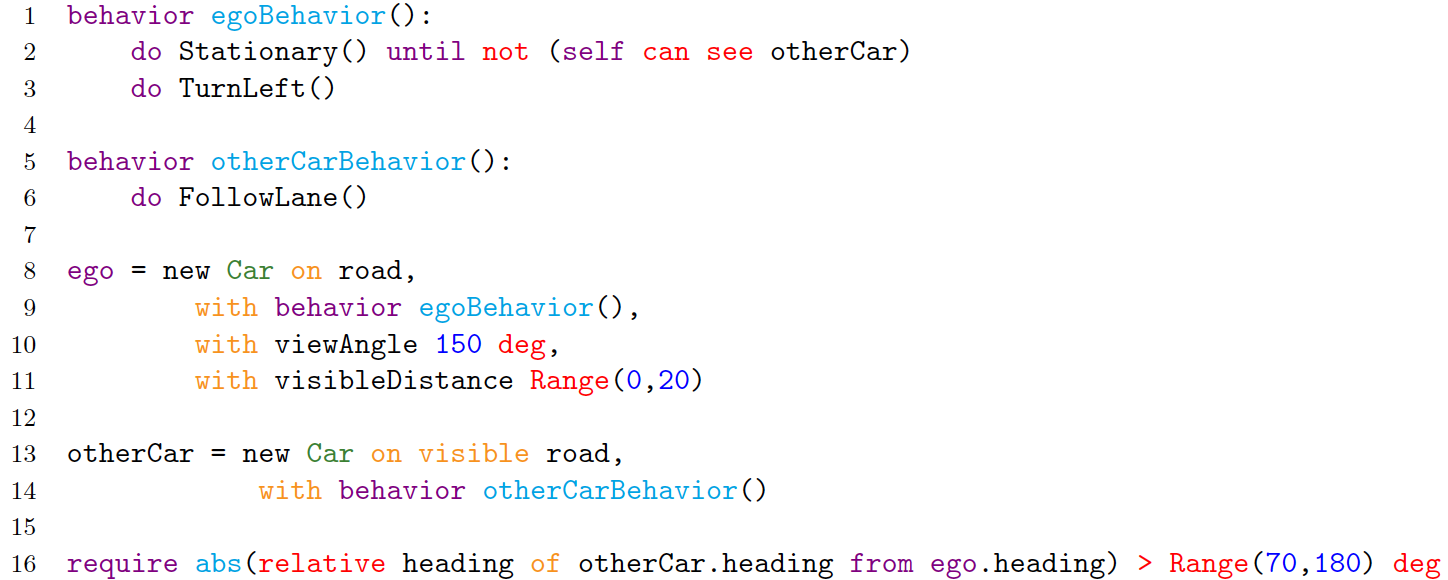}
    \caption{Accuracy Experiment (Scenario \#2)}
    \label{fig:scenario2}
\end{figure}

\begin{figure}
    \centering
    \includegraphics[width=0.8\linewidth]{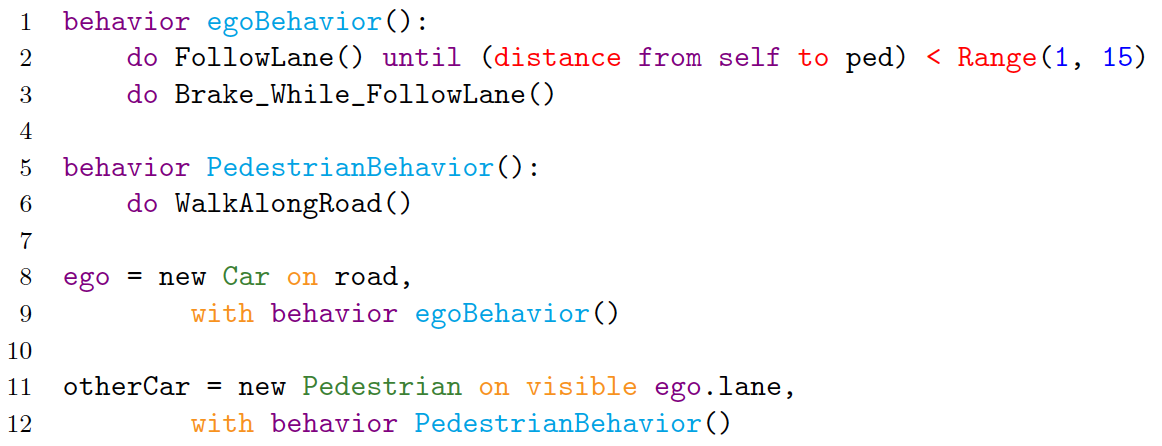}
    \caption{Accuracy Experiment (Scenario \#3)}
    \label{fig:scenario3}
\end{figure}

\begin{figure}
    \centering
    \includegraphics[width=0.8\linewidth]{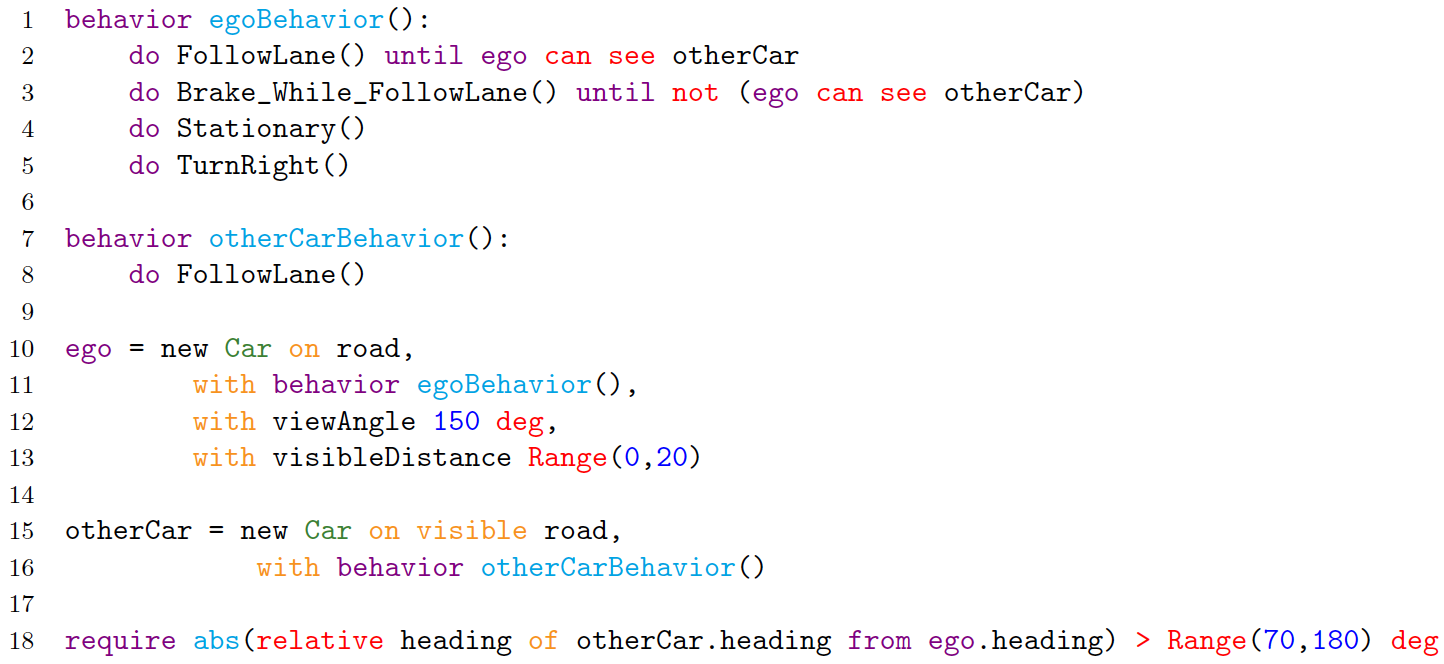}
    \caption{Accuracy Experiment (Scenario \#4)}
    \label{fig:scenario4}
\end{figure}

The four \scenic{} programs that we manually encoded for querying with our algorithm are shown in Figure~\ref{fig:scenario1}, \ref{fig:scenario2}, \ref{fig:scenario3}, \ref{fig:scenario4}.
\subsection{Prompt used for Vision Language Models (VLMs)}
We provide the following prompt for the VLMs for our experiment. \\
\textbf{Prompt:} ``You are given a sequence of images that represent traffic that are taken from a camera mounted on what we call an ego car. Please watch the video and respond whether the description of the scenario that I provide you is contained within the video for at least half of the video's duration. Your answer should be definitive: either Yes, the video contains the scenario, or No, it does not. The description consists of (1) an initial scene and (2) the behavior of ego and other cars or pedestrians. The initial scene does not have to match from the beginning of the video. What I mean by video 'containing' the scenario means that there exists a frame in the video which contains the initial scene and then the subsequent consecutive frames contains the behaviors of ego and other traffic participants as described in the scenario for at least half of the video's duration. Even if the video contains the scenario but not for at least half of the video duration, then your response should be False. Note that not all the behaviors in the scenario must occur for the scenario to be contained in the video as long as the conditions required for switching behaviors do not occur. For example, if the scenario specifies that the ego car initially follows its lane until there is a car in front on its lane with in a certain distance, and then makes a lane change. As long as there is no car in front on ego's lane, then the video of ego just following the lane for more than half of video's duration contains the scenario because the condition for making a lane change for the ego never occurred. Thus, by default, ego should be following lane. If the described scenario is contained in the video, then do the following: Per object in the description, return a frame index (0th index frame means the first image, third index frame means fourth image, etc.) from the sequence of images that I provided that contains the object and provide a description as to which object in the provided scenario description this is. Ideally, the image you return should have the full view of the object with no occlusion. However, if such an image does not exist in the video, then ensure the image contains at least half of the object. The scenario and image will be added in this prompt after this sentence. Please make sure you do everything that is asked.''

\subsection{Scenario Descriptions to the VLMs}
\textbf{Scenario\#1.} Initially, ego is on a road and there is a stationary car on ego's lane that is visible from the ego car. Ego car follows a lane by default. If the stationary car is in front of ego within 1 to 15 meters, then the ego car changes the lane. 

\textbf{Scenario\#2.} Initially, ego is on a road and there is another car that is visible to ego and coming from a different traffic direction, facing 70 to 180 degrees relative to ego's orientation. The other car follows lane. Ego is initially stationary until it cannot see the other car within 20 meters from itself in order to yield to the other car, and then it turns left. 

\textbf{Scenario\#3.} Initially, ego is on a road and there is a pedestrian on the ego's lane and is visible to ego. The pedestrian walks along the lane. Ego car follows the lane until its distance to the pedestrian is within between 1 to 15 meters. Then, ego brakes while following the lane. 

\textbf{Scenario\#4.} Initially, ego is on a road. There is another car that is visible to ego on the road heading in a different traffic direction, facing 70 to 180 degrees relative to ego's orientation. The other car follows its lane. Ego follows its lane until it sees the otherCar within 20 meters from itself. Then, it brakes while following lane until it no longer sees the car. Then, ego comes to a complete stop, being stationary and then turns right. 

\section{On Scalability Experiment}\label{appendix:scalability}
\subsection{Scalability with respect to time}
We used the \scenic{} program in Figure~\ref{fig:scenic-program} for the scalability experiment with respect to time. 

\subsection{Scalability with respect to the number of objects}
\begin{figure}
    \centering
    \includegraphics[width=0.8\linewidth]{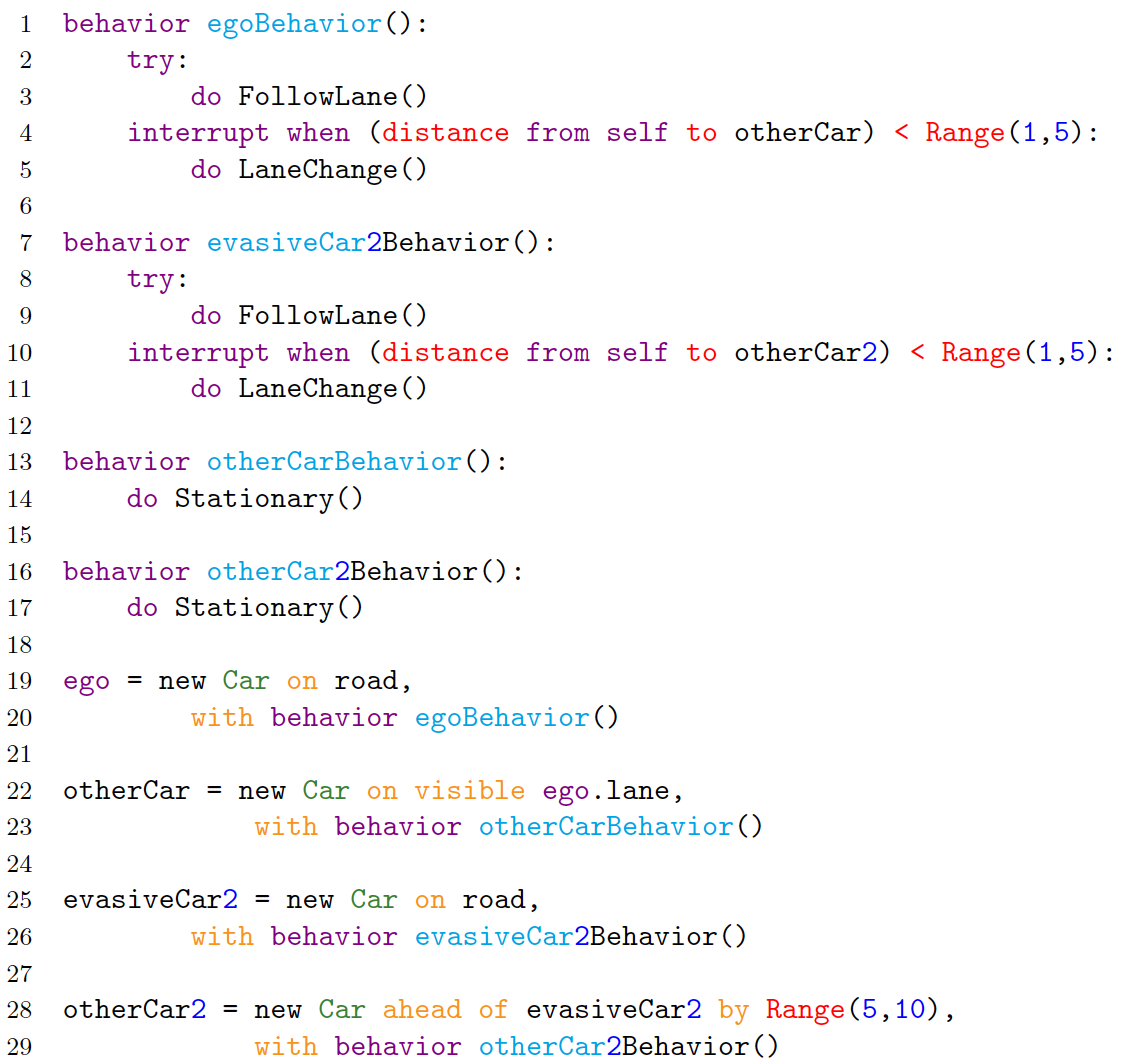}
    \caption{Scalability Experiment (Scenario with 4 objects)}
    \label{fig:scalability-4agents}
\end{figure}

We incrementally added two agents engaging in the same behavior to the same program that we used for the scalability experiment with respect to time (Figure~\ref{fig:scenic-program}). An example of a \scenic{} program with four objects are shown in Figure~\ref{fig:scalability-4agents}. Likewise, we created \scenic{} programs for six and eight objects. 

